\documentclass[review]{elsarticle}

\usepackage[utf8]{inputenc}
\usepackage{bm,graphicx,psfrag,url}
\usepackage{color}
\usepackage{amssymb}
\usepackage{caption}
\usepackage{comment}
\usepackage{graphicx}
\usepackage{wrapfig}
\usepackage{lipsum}
\usepackage{subcaption}
\usepackage{float}

\usepackage{epstopdf}

\usepackage{xcolor,tikz,colortbl}
\usepackage[cmex10]{amsmath}
\usepackage{amsthm}

\journal{Journal of \LaTeX\ Templates}

\begin{document}

\begin{frontmatter}


\title{Hierarchical Learning Using Deep Optimum-Path Forest}

\author[ufscar_addr]{Luis C. S. Afonso}
\author[unesp_addr]{Clayton R. Pereira}
\author[unesp_btu_addr]{Silke A. T. Weber}
\author[ger_addr]{Christian Hook}
\author[unicamp_addr]{Alexandre X. Falcão}
\author[unesp_addr]{João P. Papa}

\address[ufscar_addr]{UFSCar - Federal University of São Carlos, Department of Computing, São Carlos, Brazil\\  sugi.luis@ufscar.br}
\address[unesp_addr]{UNESP - São Paulo State University, School of Sciences, Bauru, Brazil\\ \{clayton.pereira,joao.papa\}@unesp.br}
\address[unesp_btu_addr]{UNESP - S\~ao Paulo State University, Medical School, Botucatu, Brazil\\ silke@fmb.unesp.br}
\address[ger_addr]{Ostbayerische Technische Hochschule, Regensburg, Germany\\ christian.hook@hs-regensburg.de}
\address[unicamp_addr]{UNICAMP - University of Campinas, Institute of Computing, Campinas, Brazil\\ afalcao@ic.unicamp.br}


\begin{abstract}
Bag-of-Visual Words (BoVW) and deep learning techniques have been widely used in several domains, which include computer-assisted medical diagnoses. In this work, we are interested in developing tools for the automatic identification of Parkinson's disease using machine learning and the concept of BoVW. The proposed approach concerns a hierarchical-based learning technique to design visual dictionaries through the Deep Optimum-Path Forest classifier. The proposed method was evaluated in six datasets derived from data collected from individuals when performing handwriting exams. Experimental results showed the potential of the technique, with robust achievements.
\end{abstract}




\end{frontmatter}



\section{Introduction}
\label{s.introduction}

Image and signal classification problems have been widely studied in the past decades by machine learning and computer vision research communities. More recently, a considerable effort is done towards deep learning (DL) techniques recently. Despite the fact that DL-driven approaches are known to be quite useful in generalizing over a number of problems, they still can not deal with some simple problems as well~\cite{NyeICLR:18}. Also, specific neural architectures need to be designed to cope with signal classification problems since most of the models available in the literature are developed to handle image-based applications only.


The well-known Bag-of-Visual-Words (BoVW)~\cite{CsurkaECCV:04} paradigm has been consistently employed and enhanced over the years to address both image- and signal-based classification problems. In a nutshell, the idea consists in extracting information (e.g., visual words/key points/descriptors) from the data for further using them to compose a dictionary (i.e., bag) that can be employed to compute new representations for a given data. Applications in medical data vary from X-ray categorization to histopathology image classification~\cite{AvniTMI:2011,CaicedoAIM:2009,SouzaSIBGRAPI:2017}, among others. 

Computer-assisted Parkinson’s disease (PD) identification is another research area that can benefit from automated diagnosis and the BoVW paradigm. Such illness is known to be neurodegenerative, it has no cure, and its main symptoms include the freezing of gate, tremors, and speech alterations, to name a few. In this context, a considerable number of works that deal with automated PD diagnosis can be referred in the literature. Spadotto et al.~\cite{SpadotoEMBC:10}, for instance, introduced the Optimum-Path Forest (OPF)~\cite{PapaIJIST:09,PapaPR:12,PapaPRL:17} for PD identification from speech signals. Later on, they employed evolutionary optimization techniques to select the most relevant features to deal with the same problem~\cite{SpadotoEMBC:11}. Sama et al.~\cite{SamaPRL:2017} and B\"{a}chlin et al.~\cite{BachlinTITB:2010} explored wearable accelerometers to detect the freezing of gate and to provide assistance as soon as the condition is detected. Rigas et al.~\cite{RigasTITB:2012} investigated an automated method that estimates the type and severity of tremors based on data acquired from accelerometers attached to specific positions at a patient's body. The estimations are used to assess both resting and action tremors.

Other works used images to cope with PD recognition automatically. Pereira et al.~\cite{Pereira:CMPB16} proposed to extract features from handwriting exams that were further digitized to fulfill the aims of the work. They used the HandPD dataset\footnote{\url{http://wwwp.fc.unesp.br/~papa/pub/datasets/Handpd}}, which comprises exams performed by healthy individuals and PD patients to detect subtle tremors when drawing spirals and meanders on a piece of paper. Since the exams were conducted using a pen equipped with sensors~\footnote{\url{https://www.oth-regensburg.de/index.php?id=5312/biometrics.html}}, the same group of authors further proposed to use the signals obtained from the pen as a means to perform automatic PD recognition~\cite{PereiraSIBGRAPI:16}. Very recently, Afonso et al.~\cite{AfonsoPR:2018} introduced the concept of ``deep recurrence plots”\ for the identification of Parkinson’s disease, where the idea is to employ recurrence plots~\cite{EckmannEPL:1987} to model the time dependency of the signals acquired during the exam.


Afonso et al.~\cite{AfonsoSIBGRAPI:17} also proposed a BoVW-based model to learn representations from signals (i.e., the same ones used in the works mentioned earlier) to be further used to cope with the problem of Parkinson’s disease identification. The proposed approach first extracts key points (descriptors) from the signal, which are then clustered using the unsupervised OPF technique~\cite{RochaIJIST:09}. The idea behind the clustering is to select only the most informative ones that will compose the final dictionary. The results showed that OPF could build more informative dictionaries than other clustering algorithms. The OPF is a framework to the design of classifiers based on graph partition, where each node stands for a dataset sample, and an adjacency relation connects them for the further application of a reward-competition approach that ends up partitioning the dataset into optimum-path trees (OPTs). Such OPTs can be either unlabeled clusters (unsupervised problems) or labeled trees (supervised and semi-supervised~\cite{AmorimPR:16} problems), and their roots stand for the so-called ``prototypes".

In this paper, we extend the work of Afonso et al.~\cite{AfonsoSIBGRAPI:17} by proposing a hierarchical-based learning methodology to design visual dictionaries. The proposed approach makes use of the Deep OPF classifier~\cite{AfonsoSIBGRAPI:16}, which aims at performing different levels of clustering to learn and encode distinct information at each phase. We showed results that outperformed the ones obtained by Afonso et al.~\cite{AfonsoSIBGRAPI:17} in the context of computer-assisted Parkinson’s disease identification using signals derived from handwriting exams.

The remainder of this paper is organized as follows. Section~\ref{s.opf} presents a theoretical background about the OPF and Deep OPF. Sections~\ref{s.repr} and~\ref{s.proposed} describe how deep representations are learned through Deep OPF and the proposed approach, respectively. Section~\ref{s.experiments} presents the experimental results, and Section~\ref{s.conclusion} states conclusions and future work.

\section{Optimum-Path Forest Clustering}
\label{s.opf}

The fundamental problem in unsupervised learning is to identify clusters in an unlabeled dataset ${\cal Z}$, such that samples from the same cluster should share some level of similarity. Many methods were proposed where the learning problem is addressed with different perspectives, such as data clustering and density estimation, just to mention a few~\cite{SchwenkerPRL:2014}. The Optimum-Path Forest handles unsupervised learning under a data clustering perspective through graph partitioning~\cite{RochaIJIST:09}. Briefly, the partition task is performed as a competitive-based process ruled by a set of key samples $\textbf{s} \in{\cal Z}$ called prototypes that conquer the remaining samples offering them optimum-cost paths. As a result, it is obtained a collection of trees (forest) rooted at each prototype, in which each tree represents a different cluster.


Suppose that a graph $(Z,{\cal A}_{k})$ can be derived from ${\cal Z}$ through a $k$-nearest neighbors adjacency relation ${\cal A}_{k}$. Each $n$-dimensional sample $\textbf{x} \in {\cal Z}$ is represented as a graph node, and the connection (edge) between two nodes $\textbf{s}$ and $\textbf{t}$ is weighted by some distance or similarity metric $d(\textbf{s},\textbf{t})$ based on their feature vectors. Also, each node $\textbf{s}$ is weighted by a probability density function (pdf) defined as follows:


\begin{equation}
\label{eq.pdf}
\rho(\textbf{s}) = \frac{1}{\sqrt{2\pi \sigma ^2} |{\cal A}(\textbf{s})|} \sum _{\forall \textbf{t}\in {\cal A}(\textbf{s})} \exp \left( \frac{-d^2(\textbf{s},\textbf{t})}{2\sigma ^2} \right),
\end{equation}
in which $\sigma = \frac{d_f}{3}$, and $d_f$ is the length of the longest edge in the graph $(Z,{\cal A}_{k})$. The choice of this parameter considers all nodes for density computation since a Gaussian function covers most samples within $d(\textbf{s},\textbf{t}) \in [0,3\sigma]$. 

The most common method for probability density function is the Parzen-window provided by Equation \ref{eq.pdf}, which is based on the isotropic Gaussian kernel when the arcs are defined by $(\textbf{s},\textbf{t}) \in {\cal A}_k$ if $d(\textbf{s},\textbf{t}) \leq d_{f}$. However, issues related to differences in scale and sample concentration arise on the application of such approach. To overcome the mentioned problems, Comaniciu~\cite{Comaniciu:03} proposed adopting adaptive choices for $d_{f}$ according to the region in the feature space. The method consists in selecting the best number of $k$-nearest neighbors within [1, $k_{max}$], such that $1\le k_{max} \le |{\cal Z}|$. In a similar way, Rocha et al. \cite{RochaIJIST:09} proposed to select the value $k \in$ [1, $k_{max}$] that minimizes the graph cut measurement of Shi and Malik~\cite{Malik:00} computed to each $(Z,{\cal A}_{k})$.


As aforementioned, the graph partitioning task is performed in a competitive fashion where the prototype nodes try to conquer the non-prototype ones by offering them optimum-path costs. A path $\pi_\textbf{t}$ in $(Z,{\cal A}_{k})$ can be defined as a sequence of adjacent nodes starting in a root-node ${\cal S}(\textbf{t})$ and ending at a sample $\textbf{t}$, where ${\cal S}$ stands for the set of root nodes. Also, let $\pi_\textbf{t}=\langle \textbf{t}\rangle$ a trivial path, and $\pi_\textbf{s}\cdot \langle \textbf{s},\textbf{t}\rangle$ the concatenation of $\pi_\textbf{s}$ and the arc $(\textbf{s},\textbf{t})$. The Optimum-Path Forest makes use of a smooth function that assigns a value $f(\pi_\textbf{t})$ to each path $\pi_\textbf{t}$. A path is said optimum if $f(\pi_\textbf{t})\ge f(\tau_\textbf{t})$, being $\tau_\textbf{t}$ any other path with terminus at $\textbf{t}$. The smooth function formulation employed by OPF is defined as follows:
 

\begin{eqnarray}
f(\langle \textbf{t} \rangle) & = & \left\{ \begin{array}{ll} 
    \rho(\textbf{t})           & \mbox{if $\textbf{t} \in {\cal S}$} \\
    \rho(\textbf{t}) - \delta  & \mbox{otherwise}
 \end{array}\right. \nonumber \\
f(\langle \pi_\textbf{s}\cdot \langle \textbf{s},\textbf{t}\rangle\rangle)&=& \min \{f(\pi_\textbf{s}), \rho(\textbf{t})\},
\label{e.pf2}
\end{eqnarray}
for $\delta = \min_{\forall (\textbf{s},\textbf{t})\in {\cal A}_k | \rho(\textbf{t}) \neq \rho(\textbf{s}) } |\rho(\textbf{t})-\rho(\textbf{s})|$. Notice that high values of delta reduce the number of maxima. In summary, the OPF algorithm maximizes $f(\pi_\textbf{t})$ such that the optimum paths form an optimum-path forest, i.e., a predecessor map $P$ with no cycles that assigns to each sample $\textbf{t}\notin {\cal S}$ its predecessor $P(\textbf{t})$ in the optimum path from ${\cal S}$ or a marker $nil$ when $\textbf{t}\in {\cal S}$.


\subsection{Deep Optimum-Path Forest}
\label{ss.dopf}


The OPF has the very interesting characteristic to identify clusters on-the-fly, which is very useful for applications where the number of groups is unknown. However, the absence of an OPF-based approach capable of computing a specific number of clusters becomes also a bottleneck. One solution is to play with the parameter $k_{max}$ by setting different values until the desired number of clusters is reached. Besides costly, this operation does not guarantee such condition is satisfied.


Based on hierarchical clustering, Afonso et al.~\cite{AfonsoSIBGRAPI:16} proposed a multi-layered OPF-based clustering algorithm. The main idea is to build a model comprised of a user-defined number of layers to obtain the desired number of clusters in the last layer. In their method, each layer is responsible for computing an optimum-path forest. The first layer takes as input the original dataset and clusters it following the OPF algorithm. The roots (prototypes) from the resulting forest are used as input by the following layer. The process is repeated until the last layer is reached. The usage of prototypes as the most representatives samples is supported by the studies of Castelo and Calder{\'o}n-Ruiz~\cite{Castelo15} and Afonso et al.~\cite{Afonso12}. Prototypes are located in the regions of highest density and, therefore, are suitable to represent the samples of its cluster~\cite{RosaICPR:2014}.

\begin{figure}[!htb]
\centerline{
  \begin{tabular}{c}
      \includegraphics[scale=0.40]{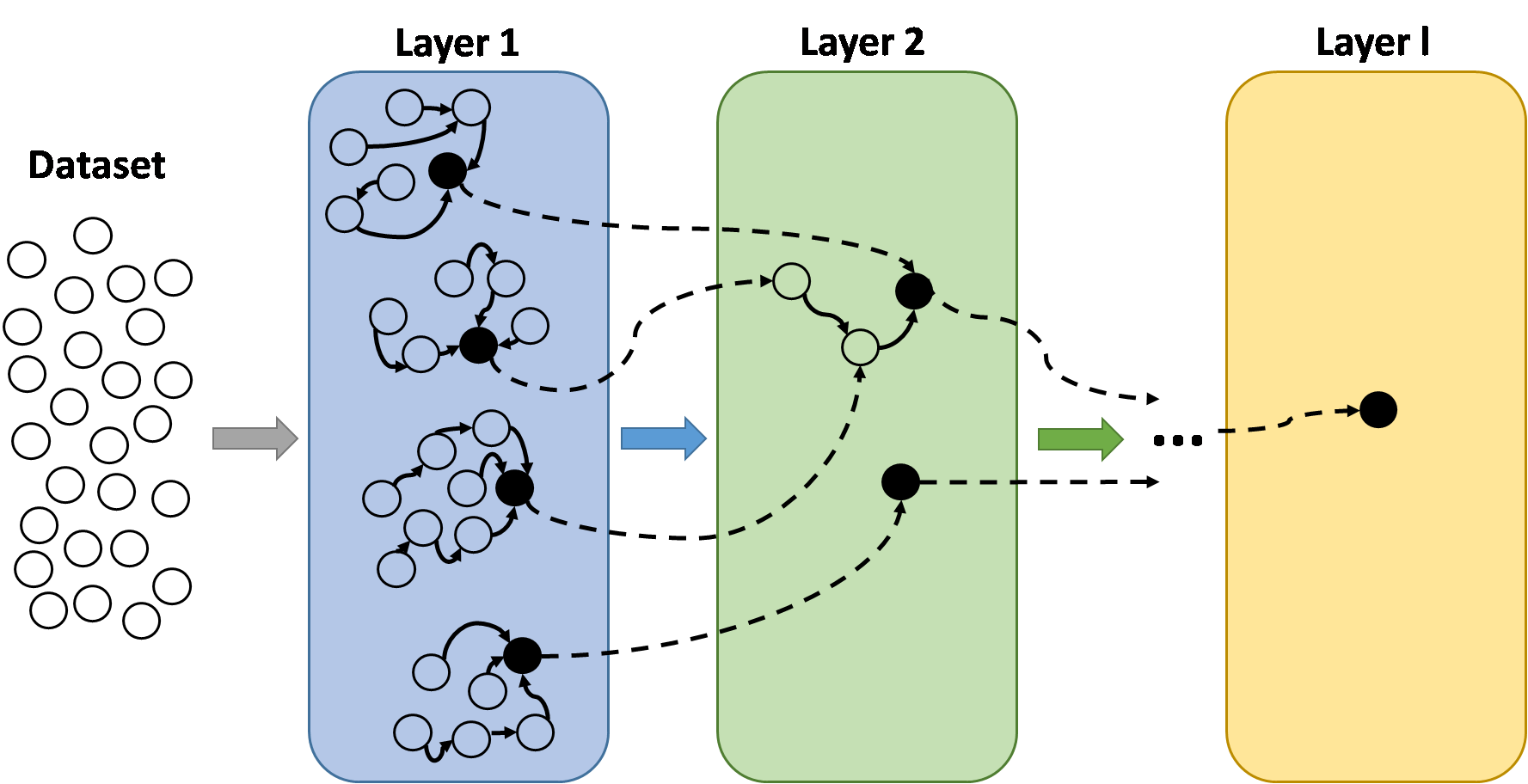}
  \end{tabular}}
  \caption{Architecture of an $l^{th}$-layered dOPF. Adapted from Afonso et al.~\cite{AfonsoSIBGRAPI:17}}
\label{f.deep_opf}
\end{figure}

Let ${\cal S}_i$ be the set of prototypes at layer $L_i$, $i=1,2,\ldots,l$, in which $l$ stands for the number of layers. Since each root will be the maximum of a pdf (Equation~\ref{eq.pdf}), we have a set of samples that fall in the same optimum-path tree and are represented by the very same prototype (root of that tree) in the next layer. In summary, the higher the number of layers, the less prototypes (clusters) one shall have, i.e., $\left|{\cal S}_1\right|<\left|{\cal S}_2\right|<\ldots<\left|{\cal S}_l\right|<\ldots\leq1$. Therefore, at layer $l$, one shall find only one cluster when $l \rightarrow \infty$. Figure~\ref{f.deep_opf} displays the OPF-based architecture for deep-driven feature space representation, hereinafter called dOPF.



In the example, the layer $L_1$ computed four clusters, i.e., optimum-path trees, rooted at the black-filled nodes (prototypes). The resulting set of prototypes ${\cal S}_1$ is used as input by the following layer $L_2$. As one can observe at layer $L_2$, a few samples become prototypes once more, resulting in the set ${\cal S}_2$. The process described above is performed until the last layer $L_l$ is reached. As the number of layers increases, the number of clusters computed by the last layer decreases, thus reducing to a single cluster at the coarsest level. This process can be interrupted as soon as the number of desired clusters (or close to it) is met.


\section{Deep-based Representations through Optimum-Path Forest}
\label{s.repr}

Deep-based representations are commonly employed in image classification applications, but they are not restricted to such ones. Such representations are obtained through deep learning architectures that are characterized by a model comprised of many layers. The introduction of such model allows learning numerous features from data as it flows through the layers. One of the most common models is the Convolutional Neural Network, which applies a series of convolutional kernels to the data, being each of them responsible for learning different information. The dOPF follows the same idea by learning multiple representations, being each of them the outcome of a clustering process from a different layer.

In the context of Bag-of-Visual Words using dOPF, the final bag could be a coarser model if only the outcome of the last layer was used to compose it (i.e., only the prototypes of the last layer comprise the bag), as proposed by Afonso et al.~\cite{AfonsoSIBGRAPI:16}. However, an enriched model could be accomplished by adding information computed by the intermediate layers as well. As a comparison, the idea of using intermediate representations would be similar to using the features learned by the many hidden layers of a deep-learning model. Each layer can learn more complex features and, therefore, more robust representations.

\begin{figure*}[htb]
\centerline{
  \begin{tabular}{c}
      \includegraphics[scale=0.37]{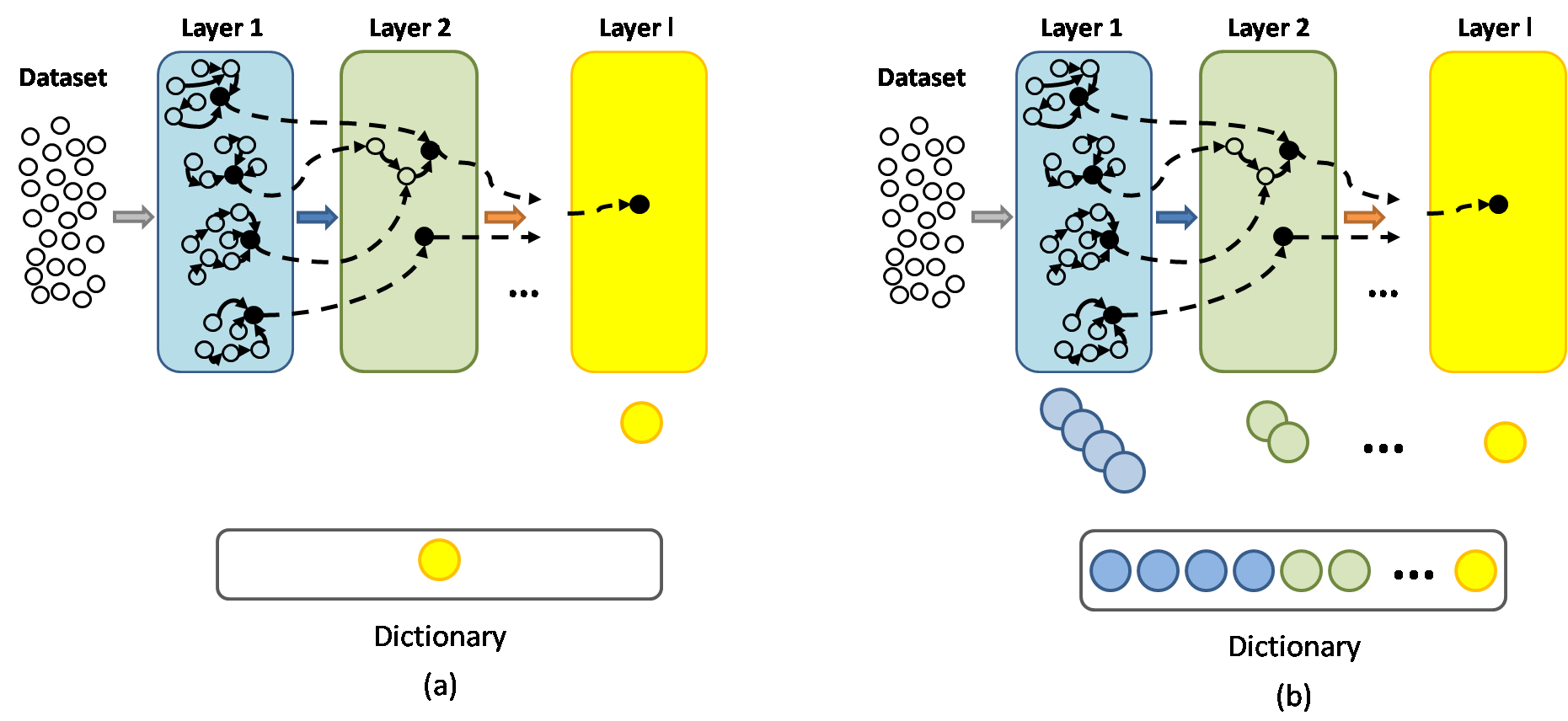}
  \end{tabular}}
  \caption{The main difference between (a) dOPF and (b) hOPF concerns the usage (or not) of features learned in the intermediate layers.}
\label{f.dhopf}
\end{figure*}

As mentioned earlier, we propose to extend the work of Afonso et al.~\cite{AfonsoSIBGRAPI:17} by employing hierarchical learning in the context of BoVW, hereinafter called hOPF (hierarchical OPF). The proposed approach will provide a more complex and more robust dictionary, being such representation the collection of selected visual words computed by all layers. Figure~\ref{f.dhopf} illustrates both dictionary learning methods, i.e., dOPF and hOPF.


Although dOPF provides a simpler and coarser representation by using only the features learned in the last layer, hOPF outputs a more complex and robust representation that stands for the concatenation of features learned by all layers. In the context of BoVW, the resulting dictionary generated by hOPF will be of size $\left|{\cal S}_1\right| + \left|{\cal S}_2\right|+\ldots+\left|{\cal S}_l\right|$.

\section{Proposed Approach}
\label{s.proposed}

\begin{figure*}[htb]
\centerline{
  \begin{tabular}{c}
      \includegraphics[scale=0.6]{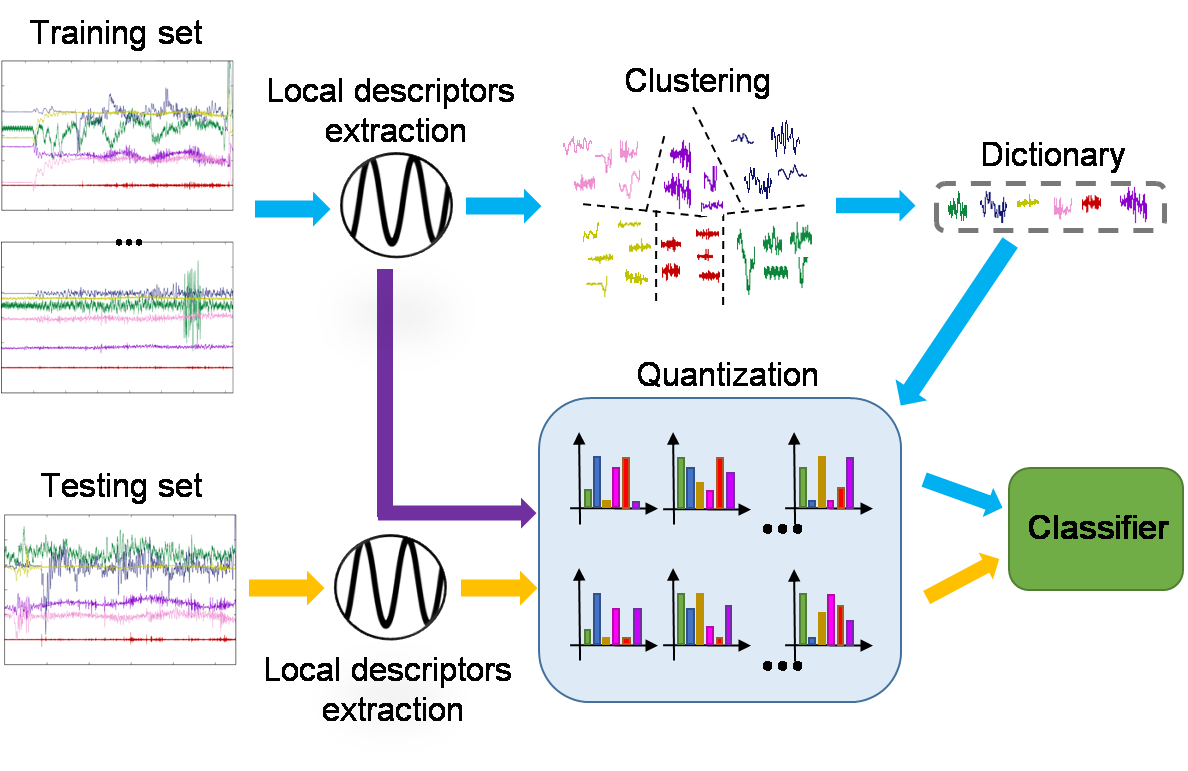}
  \end{tabular}}
  \caption{Proposed approach based on BoW and dOPF for computer-aided PD diagnosis.}
\label{f.proposed}
\end{figure*}

This section describes the steps employed in the assessment of dOPF and hOPF as visual dictionary learning methods for BoVW in the context of automatic Parkinson's disease identification, as illustrated in Figure~\ref{f.proposed}. The workflow indicated by the light blue arrow concerns the training phase. The first step computes the local descriptors from the training signals to further clustering. The most representative samples from each cluster compose the dictionary, which is used for quantization (i.e., flow indicated by the purple arrow) of both training and testing signals. The outcome of such process is the new representation of each sample. Similarly, testing signals have their local descriptors extracted and quantized (i.e., flow indicated by the yellow arrows). The final step is to perform training and classification using the new computed representations.

\subsection{Data acquisition}
\label{ss.acquisition}

The experimental data were collected from a series of tasks performed by individuals using a smart pen. The tasks exercise different hand movements that enable to capture the handwriting dynamics for further analysis. Furthermore, the exercises were elaborated in such way that are supposed not to be trivial to PD patients. All hand motion is captured by the smart pen that contains sensors that provide information on finger grip, the axial pressure of ink refill, tilt and acceleration in the $x$, $y$, and $z$ directions. 

\begin{figure}[htb]
\centerline{
  \begin{tabular}{c}
      \includegraphics[scale=0.4]{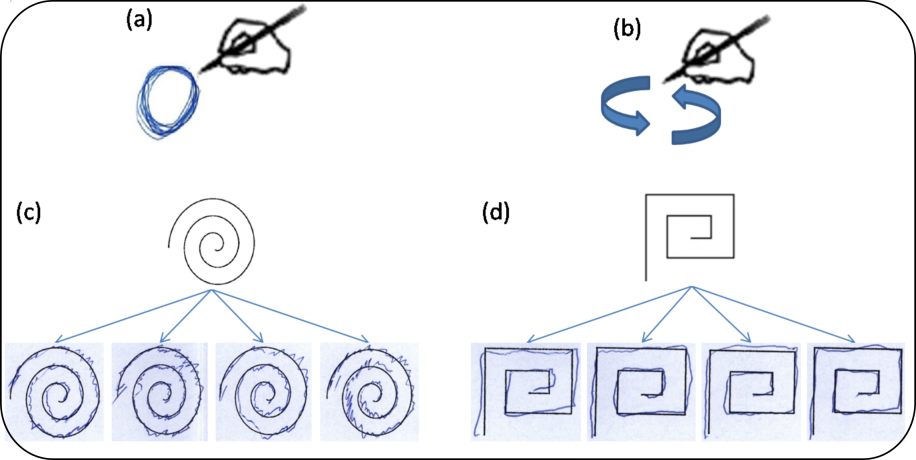}
  \end{tabular}}
  \caption{Tasks perfomed to the assessment of hand movements}
\label{f.form}
\end{figure}

Figure~\ref{f.form} illustrates the set of six tasks employed to evaluate the hand movements and to support the detection of anomalies. The set of six tasks stands for a sample, and an individual may have more than one sample assigned to it (i.e., the individual had more than on appointment). In the first task (exam (a) in Figure~\ref{f.form}), the individual is asked to draw a circle $12$ times continuously. In the second task, the individual performs the circle-drawing movement (i.e., on the air) $12$ times continuously (exam (b) in Figure~\ref{f.form}). The third and fourth tasks also concern drawing activities. In the exam (c) in Figure~\ref{f.form}, four spirals are drawn over a guideline from the inner to the outer part. The exam (d) in Figure~\ref{f.form} comprises the drawing of meander also four times and from the inner to the outer part. Last but not least, the fifth and sixth tasks are known as the diadochokinesis test and are used to evaluate the wrist movement of the right and left hands.


\subsection{Local descriptor extraction} 
\label{ss.descriptor}

\begin{figure*}[htb]
\centerline{
  \begin{tabular}{c}
      \includegraphics[scale=0.7]{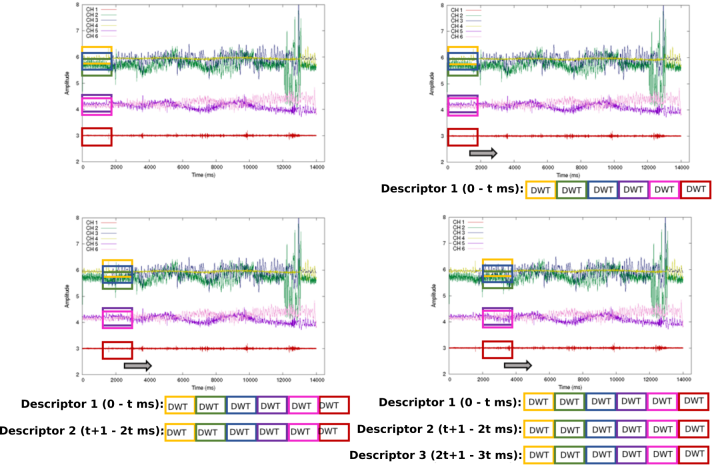}
  \end{tabular}}
  \caption{Local feature extraction.}
\label{f.dwt}
\end{figure*}

The local descriptors are extracted from the recorded signals in a sliding-window fashion that goes through each of the six signals. The descriptors are computed using a single-level Discrete Wavelet Transform (DWT) applied to each segment delimited by the sliding window. Each time segment of the signal is in fact represented by the concatenation of the resulting DWT from six sliding windows (i.e., one sliding window applied to each signal), as depicted in Figure~\ref{f.dwt}. Notice that all sliding windows comprise the same portion of time (i.e., the same initial and final times) as they go through the signals and the DWT is computed independently to each window. Moreover, the window length and shifting are user-defined. The experiments used windows of $150$ ms of length and a stride of $100$ ms, which showed the best results when compared with a window of size $100$ ms and stride of $50$ ms, and a window of size $200$ ms and stride of $150$ ms. Moreover, each segment is represented by a descriptor, and the longer is the signal, the higher will be the number of descriptors representing the input signal.

\subsection{Dictionary formulation} 
\label{ss.dictionary}

The dictionary is formulated in a straightforward way by selecting the most representative ``words" (descriptors) among the set computed in the previous step, and it is further used to compute a new sample representation. The most representative words are usually selected by a clustering algorithm where each centroid becomes a ``word"\ of the dictionary. Therefore, the dictionary size is defined by the number of clusters. Since it has some impact on the accuracy rate, it is common the use of different sizes for the dictionary to balance the computational cost and accuracy rate. As mentioned in Section~\ref{ss.dopf}, the prototypes are very suitable to represent the samples of their trees (i.e.,  prototypes are equivalent to centroids of a cluster), thus being good representations to compose the dictionaries. This work employs the Optimum-Path Forest clustering algorithm to select the most representative words, i.e., the prototypes.


\subsection{The new representation} 
\label{ss.representation}

A signal can be represented by its set of descriptors, which can range from dozens to thousands (i.e., a descriptor can be computed for each segment of the original signal, and the number of segments varies according to its length and stride). However, a few of these descriptors might be variations of another one or only represent noisy information. Moreover, machine learning techniques cannot be directly applied to the sets of descriptors since their dimension is not the same to all signals. Therefore, quantization is performed so that signals can be mapped into the same feature space. The outcome of the process is a histogram of length equals to the size of the previously formulated dictionary, where each bin stores the frequency of its closest word (descriptor) in the input signal. Finally, any machine learning technique can be applied for classification purposes using the histograms as input.


\section{Experiments and results}
\label{s.experiments}

In this section, we provide details concerning the experiments carried out on the assessment of deep-based dictionaries in the context of automatic Parkinson's disease identification. The experiments were divided into two parts: (i) the first one evaluated and compared dOPF-based dictionaries against the traditional OPF-based bags and the traditional BoVW method that computes the bags using the well-known $k$-means~(Section~\ref{ss.sibgrapi}); and (ii) the second part provides a comparison of the proposed approach (i.e., hOPF) with the method presented in the work of Afonso et al.~\cite{AfonsoSIBGRAPI:16}~(Section~\ref{ss.new}). Additionally, the second section of experiments includes the hOPF performance evaluation using compressed versions of the representations learned. For that purpose, we applied the Restricted Boltzmann Machine (RBM)~\cite{HintonNC:2002} to provide different compression levels. Notice that both experiments used data collected from $66$ exams ($35$ healthy individuals and $31$ PD patients), and the output of the protocol discussed in the previous section results in six different datasets, one for each task. The following sections describe the particularities of each experiment and the results using the proposed methodology.

\subsection{Single-scale deep-based representations}
\label{ss.sibgrapi}

This experiment aimed at evaluating the clustering quality of dOPF, $k$-means\footnote{Our implementation.} and OPF\footnote{https://github.com/jppbsi/LibOPF} through the accuracy rate obtained in the classification phase. The dOPF used in the work comprises an architecture with four layers, being the values of $k_{max}$ set as follows: $100$ for the first layer, 1\% of the number of clusters computed in the previous layer are used as an input for the second layer, and 10\% of the number of clusters computed in their respective antecessor layers for the third and fourth layers\footnote{Those values were empirically set.}. The parameter \emph{k} for $k$-means was always set as the number of clusters found by the fourth (last) layer of dOPF approach to allow a fair comparison. Regarding the OPF algorithm, the values for $k_{max}$ were empirically set as $2,500$ for the Spiral and Meander datasets, and as $1,500$ for the remaining datasets.

Table~\ref{tab.words} presents the number of local descriptors obtained from each dataset, as well as the number of visual words selected by each clustering technique. Notice the values concerning the dOPF column stand for the number of visual words selected by each layer. As aforementioned, dOPF dictionaries are comprised of the visual words computed by the last layer only (i.e., bolded values).

\begin{table}[H]	
\begin{center}
\caption{Number of descriptors extracted from the training set and number of words computed by each technique.}	
\label{tab.words}
\resizebox{\columnwidth}{!}{%
\begin{tabular}{  l | c | c | c | c }
    \textbf{dataset (task)}   	& \textbf{\# descriptors} & \textbf{dOPF}   		 	 	 & \textbf{$k$-means}   & \textbf{OPF}	  \\ \hline
    Circ-A  exam (a)        	&      18,000          	  &      5,682 - 2,584 - 228 - \textbf{68}     	 &      68        	&     693         \\
    Circ-B exam (b)        	&      11,898          	  &      538 - 376 - 43 - \textbf{17}        	 &      17        	&      33         \\
    Spiral exam (c)        	&      46,637          	  &      12,118 - 3,951 - 370 - \textbf{92}   	 &      92        	&    1,424        \\
    Meander exam (d)        	&      41,094          	  &      10,865 - 3,937 - 429 - \textbf{99} 	 &      99        	&    1,591        \\
    Dia-A exam (e)        	&      14,608          	  &      666 - 480 - 95 - \textbf{47}        	 &      47        	&       80        \\
    Dia-B exam (f)	       	&      13,947          	  &      657 - 394 - 78 - \textbf{27}        	 &      27        	&       70        \\
\end{tabular}}		    	    
\end{center}		    
\end{table}

The clustering quality was assessed under a hold-out procedure with $15$ runs, being the training and testing sets randomly partitioned in each new run and always with $50\%$ of the dataset each. The classification step employs three classifiers for comparison purposes: Bayesian Classifier (BC)\footnote{Our implementation.}, supervised OPF (sOPF)\footnote{https://github.com/LibOPF/LibOPF} and SVM using a Radial Basis Function (RBF) kernel with fine-tunned parameters (SVM-RBF)~\cite{scikit-learn}.

Tables~\ref{tab.acc.circa_all}---~\ref{tab.acc.diab_all} present the mean recognition rates concerning all six exams, being the accuracy computed according to Papa et al.~\cite{PapaIJIST:09}, which considers unbalanced datasets. The best results (i.e., bold values) are defined according to the Wilcoxon signed-rank \cite{Wilcoxon_1945} with a significance of $0.05$, which pointed out the best ones for each exam. Further, we also considered the best among all exams as the underlined ones.

The statistical evaluation pointed out [OPF, SVM-RBF] and [$k$-means, BC] as the best pairs of [dictionary learner, classifier] with accuracies near to $81\%$ and $83\%$, respectively. Comparing that recognition rates against some previous works~\cite{Pereira:CMPB16}, dOPF showed significant gains, ranging from 10\% to 30\%.

\begin{table}[H]
    \caption{Overall accuracies.}
    \begin{subtable}{0.5\linewidth}
      \centering
        \caption{Circ-A dataset.}
        \label{tab.acc.circa_all}
		\footnotesize
		\resizebox{\columnwidth}{!}{%
		\begin{tabular}{c|c|c|c}
		\textbf{}	    & \textbf{BC}			  & \textbf{sOPF}	    & \textbf{SVM-RBF}	\\ \hline
		\textbf{dOPF} 	    & \textbf{82.96$\pm$2.88} 		  & \textbf{81.71$\pm$5.12} & 73.87$\pm$4.58    \\
		\textbf{$k$-means}  & \underline{\textbf{83.38$\pm$4.22}} & \textbf{82.01$\pm$5.11} & 65.80$\pm$12.39   \\
		\textbf{OPF}        & \textbf{81.06$\pm$4.36} 		  & \textbf{81.90$\pm$4.89} & 76.17$\pm$6.92	
		\end{tabular}}
    \end{subtable} 
    \begin{subtable}{0.5\linewidth}
      \centering
        \caption{Circ-B dataset.}
        \label{tab.acc.circb_all}
		\footnotesize
		\resizebox{\columnwidth}{!}{%
		\begin{tabular}{c|c|c|c}
		\textbf{}	    & \textbf{BC}	 &	\textbf{sOPF}	   &	\textbf{SVM-RBF}	\\ \hline
		\textbf{dOPF} 	    & 68.75$\pm$7.96     & 69.14$\pm$6.95          & \textbf{77.31$\pm$4.45}	\\
		\textbf{$k$-means}  & 67.80$\pm$7.44     & 65.58$\pm$6.79          & \textbf{74.54$\pm$6.39}    \\
		\textbf{OPF}        & 70.81$\pm$4.62	 & \textbf{73.08$\pm$8.96} & \textbf{76.69$\pm$5.38}
		\end{tabular}}
    \end{subtable} \\\\
    \begin{subtable}{0.5\linewidth}
      \centering
        \caption{Spiral dataset.}
        \label{tab.acc.spiral_all}
		\footnotesize
		\resizebox{\columnwidth}{!}{%
		\begin{tabular}{c|c|c|c}
		\textbf{}	   & \textbf{BC}		& 	\textbf{sOPF}	&	\textbf{SVM-RBF}	\\ \hline
		\textbf{dOPF} 	   & \textbf{78.30$\pm$5.80}	&      76.73$\pm$6.83	&	 77.25$\pm$3.46		\\
		\textbf{$k$-means} & 73.37$\pm$5.37		&      73.11$\pm$5.31	&	 78.83$\pm$2.20		\\
		\textbf{OPF} 	   & 75.40$\pm$3.09		&      75.57$\pm$3.13	&     \textbf{81.03$\pm$2.40}		
		\end{tabular}}
    \end{subtable} 
    \begin{subtable}{0.5\linewidth}
      \centering
        \caption{Meander dataset.}
        \label{tab.acc.meander_all}
		\footnotesize
		\resizebox{\columnwidth}{!}{%
		\begin{tabular}{c|c|c|c}
		\textbf{}	  & \textbf{BC}		& \textbf{sOPF}	   & \textbf{SVM-RBF}			\\ \hline
		\textbf{dOPF} 	  & 73.33$\pm$4.97	& 74.07$\pm$2.90   & \textbf{80.45$\pm$2.42}		\\
		\textbf{$k$-means}& 76.07$\pm$3.31	& 76.09$\pm$2.77   & 78.26$\pm$3.91			\\
		\textbf{OPF} 	  & 78.53$\pm$3.15	& 77.21$\pm$3.52   & \underline{\textbf{81.07$\pm$2.60}}
		\end{tabular}}
    \end{subtable} \\\\
    \begin{subtable}{0.5\linewidth}
      \centering
        \caption{Dia-A dataset.}
        \label{tab.acc.diaa_all}
		\footnotesize
		\resizebox{\columnwidth}{!}{%
		\begin{tabular}{c|c|c|c}
		\textbf{}	   & \textbf{BC}	     & 	\textbf{sOPF}	       & \textbf{SVM-RBF}	\\ \hline
		\textbf{dOPF} 	   & \textbf{69.86$\pm$7.21} & \textbf{70.93$\pm$7.29} & \textbf{68.69$\pm$7.26}\\
		\textbf{$k$-means} & \textbf{72.18$\pm$7.46} & \textbf{72.43$\pm$5.81} & \textbf{73.93$\pm$8.66}\\
		\textbf{OPF}       & \textbf{70.72$\pm$6.60} & \textbf{67.01$\pm$7.45} & \textbf{68.69$\pm$7.26} 	
		\end{tabular}}
    \end{subtable} 
    \begin{subtable}{0.5\linewidth}
      \centering
        \caption{Dia-B dataset.}
        \label{tab.acc.diab_all}
		\footnotesize
		\resizebox{\columnwidth}{!}{%
		\begin{tabular}{c|c|c|c}
		\textbf{}	   & \textbf{BC}	     & 	\textbf{sOPF}	       & \textbf{SVM-RBF}	\\ \hline
		\textbf{dOPF} 	   & \textbf{67.96$\pm$8.10} & 64.86$\pm$7.93          & 61.89$\pm$8.49			\\
		\textbf{$k$-means} & \textbf{72.92$\pm$8.51} & \textbf{69.84$\pm$9.03} & \textbf{67.24$\pm$9.31}        \\
		\textbf{OPF}       & 63.77$\pm$8.85          & 67.25$\pm$6.80          & \textbf{66.30$\pm$7.38}
		\end{tabular}}
    \end{subtable} 
\end{table}

Concerning the best accuracies regarding each exam, dOPF obtained very much suitable results, being more accurate than na\"ive OPF in most cases. Supervised OPF obtained good results as well, but SVM-RBF achieved the best recognition rates in a few more situations. Additionally, we also evaluated the accuracy per class for all situations, as presented in Tables \ref{tab.prec.circa}---~\ref{tab.prec.diab}, whose best results are also highlighted considering the Wilcoxon signed-rank. The best results for each class are in bold, and the best among all datasets is underlined. Actually, the main improvement concerns the accuracy for the identification of healthy individuals, since Pereira et al.~\cite{PereiraSIBGRAPI:16} obtained recognition rates nearly to 50\% over the Meander and Spirals datasets for the control class. The dOPF increased not only the global accuracy with respect to the work by Pereira et al.~\cite{PereiraSIBGRAPI:16}, but also the specificity and sensitivity for most of the cases. Also, Circ-A dataset provided two out of the five best results, thus showing as a good alternative for the Parkinson's Disease identification.

    \begin{table}[H]
      \centering
	\caption{Circ-A dataset.}
	\label{tab.prec.circa}
	\resizebox{\columnwidth}{!}{%
	\begin{tabular}{c|cc|cc|cc}
	\textbf{}          & \multicolumn{2}{c|}{\textbf{BC}} 			   & \multicolumn{2}{c|}{\textbf{sOPF}} & \multicolumn{2}{c}{\textbf{SVM-RBF}} \\ \cline{2-7} 
                   	   & \textbf{HC}     		& \textbf{PD}    		      & \textbf{HC}      		    & \textbf{PD}     		& \textbf{HC}       	     & \textbf{PD}      \\ \hline
	\textbf{dOPF}      & \textbf{82.59$\pm$8.09}    & \textbf{83.33$\pm$5.62}  	      & \underline{\textbf{85.93$\pm$6.59}} & \textbf{77.5$\pm$10.24}   & \textbf{67.41$\pm$11.67}   & 61.67$\pm$15.10                 \\
	\textbf{$k$-means} & \textbf{82.59$\pm$8.09}    & \underline{\textbf{84.17$\pm$7.79}} & \textbf{84.44$\pm$12.09}            & \textbf{79.58$\pm$10.15}  & \textbf{67.04$\pm$7.99}    & 70.42$\pm$12.60                 \\
	\textbf{OPF}       & \textbf{82.96$\pm$8.52}    & \textbf{79.17$\pm$12.19}            & \textbf{82.96$\pm$6.79}             & \textbf{80.83$\pm$8.34}   & \textbf{71.48$\pm$9.82}    & 75.42$\pm$6.87                
	\end{tabular}}
    \end{table}

    \begin{table}[H]
      \centering
	\caption{Circ-B dataset.}
	\label{tab.prec.circb}
	\resizebox{\columnwidth}{!}{%
	\begin{tabular}{c|cc|cc|cc}
	\textbf{}          & \multicolumn{2}{c|}{\textbf{BC}} & \multicolumn{2}{c|}{\textbf{sOPF}} & \multicolumn{2}{c}{\textbf{SVM-RBF}} \\ \cline{2-7} 
		           & \textbf{HC}      & \textbf{PD}    		  & \textbf{HC}      & \textbf{PD}     		& \textbf{HC}       		      & \textbf{PD}               \\ \hline
	\textbf{dOPF}      & 76.67$\pm$49.12  & \textbf{60.83$\pm$18.32}  & 71.85$\pm$13.98  & \textbf{63.75$\pm$13.35} & \textbf{77.04$\pm$10.51}            & \textbf{64.58$\pm$13.24}  \\
	\textbf{$k$-means} & 77.04$\pm$12.48  & \textbf{61.25$\pm$11.68}  & 67.41$\pm$12.32  & \textbf{63.75$\pm$13.35} & \underline{\textbf{77.41$\pm$6.24}} & \textbf{68.75$\pm$14.43}  \\
	\textbf{OPF}       & 59.99$\pm$9.56   & \textbf{74.58$\pm$13.39}  & 57.04$\pm$17.15  & \textbf{54.17$\pm$13.04} & 45.93$\pm$7.98                      & \textbf{57.92$\pm$10.07}  
	\end{tabular}}
    \end{table}

    \begin{table}[H]
      \centering
	\caption{Spiral dataset.}
	\label{tab.prec.spiral}
	\resizebox{\columnwidth}{!}{%
	\begin{tabular}{c|cc|cc|cc}
	\textbf{}          & \multicolumn{2}{c|}{\textbf{BC}} & \multicolumn{2}{c|}{\textbf{sOPF}} & \multicolumn{2}{c}{\textbf{SVM-RBF}} \\ \cline{2-7} 
		           & \textbf{HC}     & \textbf{PD}    		& \textbf{HC}      & \textbf{PD}     			  & \textbf{HC}       			& \textbf{PD}		\\ \hline
	\textbf{dOPF}      & 82.08$\pm$8.14  & \textbf{74.51$\pm$10.59} & 75.42$\pm$11.92  & \underline{\textbf{78.04$\pm$12.07}} & \underline{\textbf{89.43$\pm$1.83}} & 67.81$\pm$2.17	  \\
	\textbf{$k$-means} & 81.25$\pm$11.33 & 65.49$\pm$10.63          & 79.17$\pm$10.48  & 67.06$\pm$10.14                	  & 84.85$\pm$2.25                  	& \textbf{73.59$\pm$3.67} \\
	\textbf{OPF}       & 77.90$\pm$6.95  & \textbf{72.90$\pm$6.90}  & 79.52$\pm$6.20   & \textbf{71.61$\pm$6.27}              & 86.43$\pm$1.09                  	& \textbf{74.58$\pm$0.82}
	\end{tabular}}
    \end{table} 

    \begin{table}[H]
      \centering
	\caption{Meander dataset.}
	\label{tab.prec.meander}
	\resizebox{\columnwidth}{!}{%
	\begin{tabular}{c|cc|cc|cc}
	\textbf{}          & \multicolumn{2}{c|}{\textbf{BC}} & \multicolumn{2}{c|}{\textbf{sOPF}} & \multicolumn{2}{c}{\textbf{SVM-RBF}} \\ \cline{2-7} 
		           & \textbf{HC}     & \textbf{PD}    		& \textbf{HC}      & \textbf{PD}     	     & \textbf{HC}       	& \textbf{PD}      \\ \hline
	\textbf{dOPF}      & 73.33$\pm$4.97  & \textbf{76.61$\pm$4.04}  & 72.76$\pm$5.47   & \textbf{75.38$\pm$4.62} & \textbf{85.80$\pm$0.89}  & 74.81$\pm$2.18   \\
	\textbf{$k$-means} & 82.67$\pm$4.88  & 69.46$\pm$5.96           & 78.95$\pm$4.99   & 73.23$\pm$4.51          & \textbf{84.43$\pm$3.76}  & 71.06$\pm$2.62   \\
	\textbf{OPF}       & 80.29$\pm$4.68  & \textbf{76.77$\pm$8.44}  & 76.57$\pm$5.47   & \textbf{77.85$\pm$3.93} & 87.99$\pm$0.72           & 74.54$\pm$1.37                
	\end{tabular}}
    \end{table}  

    \begin{table}[H]
      \centering
	\caption{Dia-A dataset.}
	\label{tab.prec.diaa}
	\resizebox{\columnwidth}{!}{%
	\begin{tabular}{c|cc|cc|cc}
	\textbf{}          & \multicolumn{2}{c|}{\textbf{BC}} & \multicolumn{2}{c|}{\textbf{sOPF}} & \multicolumn{2}{c}{\textbf{SVM-RBF}} \\ \cline{2-7} 
		           & \textbf{HC}     	      & \textbf{PD}    		  & \textbf{HC}      	      & \textbf{PD}     	  & \textbf{HC}       	      & \textbf{PD}      	\\ \hline
	\textbf{dOPF}      & \textbf{72.22$\pm$9.51}  & \textbf{67.50$\pm$16.01}  & \textbf{78.52$\pm$8.33}   & \textbf{65.83$\pm$13.28}  & \textbf{75.19$\pm$7.55}   & \textbf{66.25$\pm$13.46}  \\
	\textbf{$k$-means} & \textbf{75.19$\pm$5.69}  & \textbf{66.67$\pm$14.01}  & \textbf{74.44$\pm$6.97}   & \textbf{70.42$\pm$12.59}  & \textbf{67.78$\pm$12.37}  & \textbf{66.25$\pm$15.61}  \\
	\textbf{OPF}       & \textbf{50.74$\pm$9.70}  & \textbf{51.67$\pm$17.15}  &  \textbf{55.56$\pm$12.67} & \textbf{52.08$\pm$11.10}  & \textbf{50.00$\pm$13.46}  & \textbf{47.50$\pm$12.87}                
	\end{tabular}}
    \end{table} 

    \begin{table}[H]
      \centering
	\caption{Dia-B dataset.}
	\label{tab.prec.diab}
	\resizebox{\columnwidth}{!}{%
	\begin{tabular}{c|cc|cc|cc}
	\textbf{}          & \multicolumn{2}{c|}{\textbf{BC}} & \multicolumn{2}{c|}{\textbf{sOPF}} & \multicolumn{2}{c}{\textbf{SVM-RBF}} \\ \cline{2-7} 
		           & \textbf{HC}     	      & \textbf{PD}    		 & \textbf{HC}      	     & \textbf{PD}     	         & \textbf{HC}       	     & \textbf{PD}      \\ \hline
	\textbf{dOPF}      & \textbf{72.59$\pm$9.62}  & \textbf{63.33$\pm$12.88} & \textbf{73.33$\pm$9.56}   & \textbf{72.50$\pm$10.89}  & \textbf{56.29$\pm$15.56}  & \textbf{71.25$\pm$11.81}  \\
	\textbf{$k$-means} & \textbf{68.89$\pm$9.89}  & 60.83$\pm$11.29          & \textbf{72.59$\pm$10.63}  & \textbf{67.08$\pm$13.39}  & 60.74$\pm$11.38           & \textbf{73.75$\pm$11.23}  \\
	\textbf{OPF}       & 50.37$\pm$28.07          & 60.00$\pm$32.18          & 48.52$\pm$12.42           & 47.92$\pm$10.62           & \textbf{53.70$\pm$9.44}   & \textbf{52.08$\pm$15.92}                
	\end{tabular}}
    \end{table}

Table~\ref{tab.time} presents the mean computational load required by each technique for dictionary learning. Notice the computational burden for dOPF considers the four layers. In this context, $k$-means figured as the fastest one due to its simplicity. If one considers dOPF and OPF only, we can observe the former is about $78$ times faster in Circ-B dataset, which is quite effective. The lowest gains can be observed in both Meander and Spiral datasets. The small differences come from the fact the value used for $k_{max}$ in both situations is small, thus justifying the fact the dictionaries computed in these datasets have very high dimension when compared to others.

\begin{table}[H]	
\begin{center}
\caption{Dictionary learning computational load [s] required by each technique.}
\label{tab.time}	
\begin{tabular}{  l | c | c | c }
    \textbf{dataset (task)}   	& \textbf{dOPF}   	& \textbf{$k$-means}   	& \textbf{OPF}	  \\ \hline
    Circ-A  (a)        		&	  968.167	&	 37.008		&	49,087.137	\\
    Circ-B  (b)        		&	  419.498	&	 13.113		&	32,777.539	\\
    Spiral  (c)        		&	6,063.205	&	239.859		&	 6,643.906	\\
    Meander (d)        		&	5,003.233	&	208.443		&	 5,168.819	\\
    Dia-A   (e)        		&	  613.109	&	 19.878		&	41,189.133	\\
    Dia-B   (f)	       		&	  569.053	&	 11.025		&	39,367.844	\\
\end{tabular}		    	    
\end{center}		    
\end{table}

\subsection{Multi-scale deep-based representations}
\label{ss.new}

\begin{figure}[H]
\centerline{
  \begin{tabular}{c}
      \includegraphics[scale=0.55]{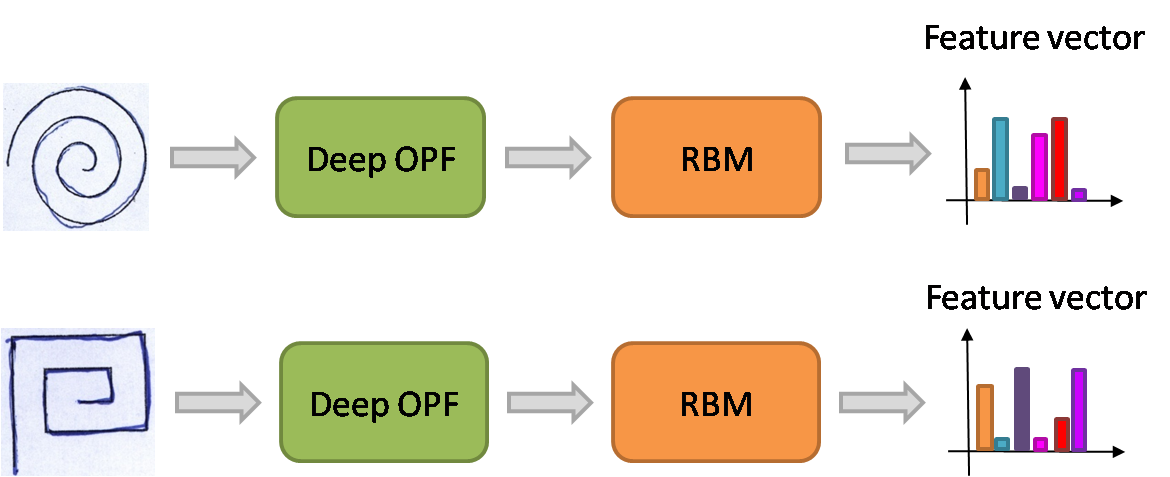}
  \end{tabular}}
  \caption{The Deep OPF block represents the workflow depicted in Figure~\ref{f.proposed}. The outcome of such process is used as input of RBM that outputs a compressed representation used by classifiers.}
\label{f.rbm}
\end{figure}

As aforementioned, this round of experiments aimed at providing a performance comparison between dOPF and hOPF. To fulfill that purpose, the quality of the dictionaries provided by both techniques was compared using the protocol described in Section~\ref{ss.sibgrapi}. As more visual words were added, hOPF dictionaries provide higher-dimensional representations. Hence, an additional experiment evaluated the quality of compressed representations computed by RBM. There were used representation sizes of $25\%$ (hOPF-25), $50\%$ (hOPF-50), and $75\%$ (hOPF-75) of the original one (hOPF). Figure~\ref{f.rbm} illustrates the workflow of representation compression.

Tables~\ref{tab.dh.acc.circa_all}--~\ref{tab.dh.acc.diab_all} provide the overall accuracy rates concerning each dataset. The accuracy rate was computed using the same formulation as in Section~\ref{ss.sibgrapi}, and the best results (i.e., bolded ones) were determined by the Wilcoxon signed-rank with significance as $0.05$. 

In general, hOPF-based dictionaries achieved competitive results in all six datasets and always figured among the best ones. Also, slight improvements compared to dOPF can be observed in most scenarios, being the most significant ones achieved in the Dia-B dataset (Table~\ref{tab.dh.acc.diab_all}). The average gain in that dataset varies from $5.79\%$ (BC) to $8.83\%$ (SVM-RBF).

An interesting aspect to be highlighted, it is the fact that compressed representations computed by RBM also figured among the best results, even the most compressed ones (hOPF-25). The representations hOPF-50 and hOPF-75 achieved the best performance among the compressed versions with best results in $11$ out of $18$ scenarios against $6$ out $18$ of hOPF-25.

Concerning the classifiers employed in the work, it can be observed a similar situation like the one illustrated in Section~\ref{ss.sibgrapi}. The classifiers obtained good results with SVM-RBF being the best statistically technique in most cases. The highest accuracy among all datasets was achieved by the pair [hOPF, SVM-RBF] with $85.29\%$.

\begin{table}[H]
\centering
        \caption{Circ-A dataset - Overall accuracies.}
        \label{tab.dh.acc.circa_all}
		\footnotesize
		\resizebox{\columnwidth}{!}{%
		\begin{tabular}{c|c|c|c|c|c}
		\textbf{}		& \textbf{dOPF}	  	  & \textbf{hOPF}  	     & \textbf{hOPF-25}  	 & \textbf{hOPF-50}        & \textbf{hOPF-75}	\\ \hline
		\textbf{BC} 		& \textbf{82.94$\pm$6.00} & \textbf{82.94$\pm$5.69}  & \textbf{79.80$\pm$6.75}   & \textbf{81.96$\pm$7.53}  & \textbf{80.20$\pm$6.14}	\\
		\textbf{sOPF}		& \textbf{82.16$\pm$6.43} & \textbf{82.94$\pm$5.69}  & \textbf{79.22$\pm$6.71}   & \textbf{80.39$\pm$7.18}  & \textbf{79.80$\pm$6.14}	\\
		\textbf{SVM-RBF}      	& \textbf{84.51$\pm$5.82} & \textbf{85.29$\pm$3.69}  & \textbf{83.14$\pm$4.19}   & \textbf{84.12$\pm$8.23}  & \textbf{81.76$\pm$4.87}	
		\end{tabular}}
\end{table}

\begin{table}[H]
\centering
        \caption{Circ-B dataset - Overall accuracies.}
        \label{tab.dh.acc.circb_all}
		\footnotesize
		\resizebox{\columnwidth}{!}{%
		\begin{tabular}{c|c|c|c|c|c}
		\textbf{}		& \textbf{dOPF}	 	  & \textbf{hOPF}  	     & \textbf{hOPF-25}  	& \textbf{hOPF-50}  	   & \textbf{hOPF-75}	\\ \hline
		\textbf{BC} 		& \textbf{80.98$\pm$3.49} & \textbf{79.22$\pm$4.65}  & 69.80$\pm$5.50     	& 68.63$\pm$8.73  	   & 68.04$\pm$8.67	\\
		\textbf{sOPF}		& \textbf{80.00$\pm$3.88} & \textbf{79.02$\pm$5.87}  & 69.61$\pm$5.63     	& 68.63$\pm$8.73  	   & 68.04$\pm$8.81	\\
		\textbf{SVM-RBF}      	& \textbf{79.61$\pm$6.33} & \textbf{79.02$\pm$4.57}  & \textbf{78.24$\pm$6.07}  & \textbf{78.04$\pm$5.76}  & \textbf{79.41$\pm$7.94}
		\end{tabular}}
\end{table}

\begin{table}[H]
\centering
        \caption{Spiral dataset - Overall accuracies.}
        \label{tab.dh.acc.spiral_all}
		\footnotesize
		\resizebox{\columnwidth}{!}{%
		\begin{tabular}{c|c|c|c|c|c}
		\textbf{}		& \textbf{dOPF}	 	  & \textbf{hOPF}  	     & \textbf{hOPF-25}  	& \textbf{hOPF-50}  	   & \textbf{hOPF-75}	\\ \hline
		\textbf{BC} 		& \textbf{77.22$\pm$3.57} & \textbf{78.94$\pm$2.58}  & 74.04$\pm$3.64     	& 72.88$\pm$2.35  	   & 72.32$\pm$3.36	\\
		\textbf{sOPF}		& \textbf{76.97$\pm$3.64} & \textbf{77.88$\pm$2.59}  & 73.33$\pm$3.64     	& 72.83$\pm$2.35  	   & 71.77$\pm$3.36	\\
		\textbf{SVM-RBF}      	& \textbf{79.49$\pm$2.33} & \textbf{81.21$\pm$2.13}  & \textbf{80.35$\pm$1.44}  & \textbf{80.15$\pm$2.73}  & \textbf{80.91$\pm$2.61}		
		\end{tabular}}
\end{table}

\begin{table}[H]
\centering
        \caption{Meander dataset - Overall accuracies.}
        \label{tab.dh.acc.meander_all}
		\footnotesize
		\resizebox{\columnwidth}{!}{%
		\begin{tabular}{c|c|c|c|c|c}
		\textbf{}		& \textbf{dOPF}	 	    & \textbf{hOPF}  	       & \textbf{hOPF-25}  & \textbf{hOPF-50} & \textbf{hOPF-75}	\\ \hline
		\textbf{BC} 		& \textbf{77.02$\pm$3.39}   & \textbf{78.64$\pm$3.05}  & 68.89$\pm$4.59     & 68.79$\pm$3.92  	& 68.38$\pm$4.26	\\
		\textbf{sOPF}		& 75.15$\pm$ 3.09	    & \textbf{77.42$\pm$3.29}  & 68.89$\pm$4.37     & 68.28$\pm$4.13  	& 67.68$\pm$3.66	\\
		\textbf{SVM-RBF}      	& \textbf{82.17$\pm$3.82}   & \textbf{83.79$\pm$2.51}  & 79.04$\pm$2.21     & 79.55$\pm$2.18  	& 78.74$\pm$3.63
		\end{tabular}}
\end{table}

\begin{table}[H]
\centering
        \caption{Dia-A dataset - Overall accuracies.}
        \label{tab.dh.acc.diaa_all}
		\footnotesize
		\resizebox{\columnwidth}{!}{%
		\begin{tabular}{c|c|c|c|c|c}
		\textbf{}		& \textbf{dOPF}	      	  & \textbf{hOPF}           & \textbf{hOPF-25}       & \textbf{hOPF-50}    	  & \textbf{hOPF-75}	\\ \hline
		\textbf{BC} 		& \textbf{73.33$\pm$7.90} & \textbf{73.33$\pm$4.95}  & 66.47$\pm$6.53          & \textbf{69.22$\pm$7.28}  & \textbf{69.99$\pm$8.63}	\\
		\textbf{sOPF}		& \textbf{73.53$\pm$7.86} & \textbf{73.33$\pm$5.83}  & 66.47$\pm$6.92          & \textbf{68.24$\pm$7.31}  & \textbf{68.82$\pm$7.36}	\\
		\textbf{SVM-RBF}      	& \textbf{79.22$\pm$5.38} & \textbf{77.25$\pm$4.51}  & \textbf{76.86$\pm$4.56} & \textbf{75.69$\pm$4.37}  & \textbf{79.61$\pm$7.49}
		\end{tabular}}
\end{table}

\begin{table}[H]
\centering
        \caption{Dia-B dataset - Overall accuracies.}
        \label{tab.dh.acc.diab_all}
		\footnotesize
		\resizebox{\columnwidth}{!}{%
		\begin{tabular}{c|c|c|c|c|c}
		\textbf{}		& \textbf{dOPF}	 & \textbf{hOPF}  & \textbf{hOPF-25}  & \textbf{hOPF-50}  & \textbf{hOPF-75}	\\ \hline
		\textbf{BC} 		& 68.43$\pm$5.71 & 68.43$\pm$5.72  & 69.80$\pm$6.53     & \textbf{75.29$\pm$5.54}  & \textbf{73.14$\pm$7.45}	\\
		\textbf{sOPF}		& 67.45$\pm$6.43 & 67.45$\pm$6.43  & 69.41$\pm$6.65     & \textbf{74.71$\pm$6.27}  & \textbf{73.14$\pm$7.02}	\\
		\textbf{SVM-RBF}      	& 70.39$\pm$6.43 & 74.51$\pm$5.52  & 74.31$\pm$9.06     & \textbf{77.84$\pm$8.60}  & \textbf{80.59$\pm$5.07}
		\end{tabular}}
\end{table}

We also investigated the accuracy rates in each class, as shown in Tables~\ref{tab.dh.acc.circa_classes}--~\ref{tab.dh.acc.diab_classes}. The representations learned by the hierarchical approach also figured among the best results in many situations. Once again, the more significant improvements (i.e., compared to dOPF) can be observed in the HC class in almost all scenarios, such as the ones in Circ-B and Dia-B datasets (i.e., the greatest ones). Compressed representations also presented competitive results, especially the most compressed one (hOPF-25) as one can observe in Circ-A and Circ-B datasets for HC class, and Dia-A dataset for PD class.

\begin{table}[H]
\centering
\caption{Circ-A - Average accuracy rate for each class.}
\label{tab.dh.acc.circa_classes}
\resizebox{\columnwidth}{!}{%
\begin{tabular}{c|cc|cc|cc}
\textbf{}         & \multicolumn{2}{c|}{\textbf{BC}} & \multicolumn{2}{c|}{\textbf{sOPF}} & \multicolumn{2}{c}{\textbf{SVM-RBF}} \\ \cline{2-7} 
                  & \textbf{HC}     	     & \textbf{PD}    		& \textbf{HC}      	    & \textbf{PD}     		& \textbf{HC}              & \textbf{PD}      	\\ \hline
\textbf{dOPF}     & \textbf{81.85$\pm$10.17} & \textbf{84.17$\pm$13.34} & \textbf{81.85$\pm$10.17}  & \textbf{82.50$\pm$12.32}  & \textbf{80.74$\pm$11.85} & \textbf{88.75$\pm$11.62}            \\
\textbf{hOPF}    & 77.78$\pm$8.91           & \textbf{85.83$\pm$11.68} & 78.52$\pm$8.62            & \textbf{86.25$\pm$12.54}  & 77.41$\pm$8.26           & \textbf{88.75$\pm$13.81}            \\
\textbf{hOPF-25} & \textbf{85.56$\pm$8.08}  & 73.33$\pm$14.07          & \textbf{85.93$\pm$8.36}   & 71.67$\pm$13.75           & \textbf{80.74$\pm$12.04}  & \textbf{85.83$\pm$10.94}            \\
\textbf{hOPF-50} & \textbf{85.19$\pm$7.76}  & \textbf{78.33$\pm$13.12} & \textbf{84.81$\pm$7.41}   & 75.42$\pm$11.68           & \textbf{85.56$\pm$7.21}  & \textbf{82.50$\pm$14.98}            \\
\textbf{hOPF-75} & \textbf{84.44$\pm$7.33}  & 75.42$\pm$8.34           & \textbf{84.44$\pm$7.33}   & 74.58$\pm$8.67            & \textbf{84.44$\pm$7.62}   & 78.75$\pm$11.76                            
\end{tabular}}
\end{table}

\begin{table}[H]
\centering
\caption{Circ-B - Average accuracy rate for each class.}
\label{tab.dh.acc.circb_classes}
\resizebox{\columnwidth}{!}{%
\begin{tabular}{c|cc|cc|cc}
\textbf{}         & \multicolumn{2}{c|}{\textbf{BC}} & \multicolumn{2}{c|}{\textbf{sOPF}} & \multicolumn{2}{c}{\textbf{SVM-RBF}} \\ \cline{2-7} 
                  & \textbf{HC}     	     & \textbf{PD}    		& \textbf{HC}      	    & \textbf{PD}     	       & \textbf{HC}       	    & \textbf{PD}      \\ \hline
\textbf{dOPF}     & 78.52$\pm$9.12           & \textbf{83.75$\pm$9.97}  & 78.15$\pm$9.02            & \textbf{82.08$\pm$9.99}  & \textbf{79.63$\pm$11.04}   & \textbf{79.58$\pm$12.38}  \\
\textbf{hOPF}    & 78.52$\pm$6.92           & \textbf{80.00$\pm$7.17}  & 77.41$\pm$8.26            & \textbf{80.83$\pm$9.59}  & \textbf{80.74$\pm$9.12}    & \textbf{77.08$\pm$8.41}   \\
\textbf{hOPF-25} & \textbf{86.67$\pm$5.86}  & 50.83$\pm$12.47          & \textbf{86.67$\pm$5.86}   & 50.42$\pm$12.60          & \textbf{86.30$\pm$10.47}   & 69.17$\pm$8.34            \\
\textbf{hOPF-50} & \textbf{85.56$\pm$6.90}  & 49.58$\pm$16.61          & \textbf{85.93$\pm$7.23}   & 49.17$\pm$16.68          & \textbf{81.85$\pm$9.26}    & \textbf{73.75$\pm$14.02}  \\
\textbf{hOPF-75} & 78.15$\pm$13.36          & 56.67$\pm$14.84          & \textbf{78.15$\pm$14.77}  & 56.67$\pm$14.26          & \textbf{86.67$\pm$8.08}    & \textbf{71.25s$\pm$16.33}
\end{tabular}}
\end{table}

\begin{table}[H]
\centering
\caption{Spiral - Average accuracy rate for each class.}
\label{tab.dh.acc.spiral_classes}
\resizebox{\columnwidth}{!}{%
\begin{tabular}{c|cc|cc|cc}
\textbf{}         & \multicolumn{2}{c|}{\textbf{BC}} & \multicolumn{2}{c|}{\textbf{sOPF}} & \multicolumn{2}{c}{\textbf{SVM-RBF}} \\ \cline{2-7} 
                  & \textbf{HC}     	     & \textbf{PD}    	       & \textbf{HC}     	  & \textbf{PD}     	       & \textbf{HC}               & \textbf{PD}      \\ \hline
\textbf{dOPF}     & 78.76$\pm$6.91           & \textbf{75.48$\pm$5.89} & 78.19$\pm$6.71           & \textbf{75.59$\pm$6.09}    & \textbf{84.95$\pm$5.91}   & \textbf{73.33$\pm$5.91}   \\
\textbf{hOPF}    & \textbf{83.24$\pm$3.48}  & \textbf{74.09$\pm$4.42} & \textbf{82.38$\pm$4.17}  & \textbf{72.80$\pm$4.35}    & \textbf{86.38$\pm$5.79}   & \textbf{75.38$\pm$6.72}   \\
\textbf{hOPF-25} & \textbf{84.95$\pm$4.07}  & 61.72$\pm$6.75          & \textbf{84.38$\pm$3.98}  & 60.86$\pm$6.66             & \textbf{87.14$\pm$6.01}   & \textbf{72.69$\pm$6.17}   \\
\textbf{hOPF-50} & 81.90$\pm$4.06           & 62.69$\pm$5.13          & 82.10$\pm$3.58           & 62.37$\pm$4.79             & \textbf{84.10$\pm$4.61}   & \textbf{75.70$\pm$4.46}   \\
\textbf{hOPF-75} & \textbf{82.67$\pm$4.48}  & 60.65$\pm$6.86          & \textbf{82.19$\pm$4.58}  & 60.00$\pm$6.93             & \textbf{87.52$\pm$6.33}   & \textbf{73.44$\pm$6.06}
\end{tabular}}
\end{table}

\begin{table}[H]
\centering
\caption{Meander - Average accuracy rate for each class.}
\label{tab.dh.acc.meander_classes}
\resizebox{\columnwidth}{!}{%
\begin{tabular}{c|cc|cc|cc}
\textbf{}         & \multicolumn{2}{c|}{\textbf{BC}} & \multicolumn{2}{c|}{\textbf{sOPF}} & \multicolumn{2}{c}{\textbf{SVM-RBF}} \\ \cline{2-7} 
                  & \textbf{HC}     	     & \textbf{PD}    	       & \textbf{HC}     	   & \textbf{PD}     	      & \textbf{HC}       	   & \textbf{PD}      \\ \hline
\textbf{dOPF}     & \textbf{80.76$\pm$5.50}  & \textbf{72.80$\pm$7.00} & \textbf{77.43$\pm$7.56}   & \textbf{72.58$\pm$7.44}  & 89.43$\pm$4.41             & \textbf{73.98$\pm$7.41}   \\
\textbf{hOPF}    & \textbf{82.19$\pm$5.29}  & \textbf{74.62$\pm$3.19} & \textbf{80.10$\pm$5.83}   & \textbf{74.41$\pm$3.49}  & \textbf{92.48$\pm$3.98}    & \textbf{73.98$\pm$3.33}   \\
\textbf{hOPF-25} & 71.14$\pm$7.41           & 66.34$\pm$6.87          & 71.14$\pm$7.21            & 66.34$\pm$7.00           & \textbf{90.38$\pm$4.09}    & 66.24$\pm$3.93            \\
\textbf{hOPF-50} & 72.86$\pm$6.39           & 64.19$\pm$6.63          & 72.10$\pm$6.48            & 63.98$\pm$6.81           & \textbf{90.76$\pm$4.58}    & 66.88$\pm$4.35            \\
\textbf{hOPF-75} & 72.48$\pm$5.49           & 63.76$\pm$4.60          & 71.14$\pm$4.52            & 63.76$\pm$4.60           & \textbf{90.29$\pm$5.92}    & 65.70$\pm$5.08                            
\end{tabular}}
\end{table}

\begin{table}[H]
\centering
\caption{Dia-A - Average accuracy rate for each class.}
\label{tab.dh.acc.diaa_classes}
\resizebox{\columnwidth}{!}{%
\begin{tabular}{c|cc|cc|cc}
\textbf{}         & \multicolumn{2}{c|}{\textbf{BC}} & \multicolumn{2}{c|}{\textbf{sOPF}} & \multicolumn{2}{c}{\textbf{SVM-RBF}} \\ \cline{2-7} 
                  & \textbf{HC}     	      & \textbf{PD}    	         & \textbf{HC}      	     & \textbf{PD}     	         & \textbf{HC}                & \textbf{PD}      \\ \hline
\textbf{dOPF}     & \textbf{79.26$\pm$7.99}   & \textbf{66.67$\pm$13.08} & \textbf{78.52$\pm$6.60}   & \textbf{67.92$\pm$12.47}  & \textbf{72.96$\pm$9.59}    & \textbf{86.25$\pm$10.88}   \\
\textbf{hOPF}    & 72.96$\pm$8.19            & \textbf{73.75$\pm$13.04} & 72.96$\pm$8.62            & \textbf{73.75$\pm$14.98}  & 72.22$\pm$7.86             & \textbf{82.92$\pm$13.04}   \\
\textbf{hOPF-25} & 64.81$\pm$6.79            & \textbf{68.33$\pm$15.13} & 64.81$\pm$7.76            & \textbf{68.33$\pm$14.46}  & \textbf{73.70$\pm$9.02}    & \textbf{80.42$\pm$10.79}   \\
\textbf{hOPF-50} & 72.59$\pm$10.59           & \textbf{65.42$\pm$14.15} & \textbf{70.74$\pm$10.60}  & \textbf{65.42$\pm$14.54}  & \textbf{80.37$\pm$9.82}    & 70.42$\pm$11.68            \\
\textbf{hOPF-75} & 67.41$\pm$12.58           & \textbf{72.92$\pm$19.00} & 65.19$\pm$12.68           & \textbf{72.92$\pm$18.85}  & \textbf{76.67$\pm$10.33}   & \textbf{82.92$\pm$14.07}
\end{tabular}}
\end{table}

\begin{table}[H]
\centering
\caption{Dia-B - Average accuracy rate for each class.}
\label{tab.dh.acc.diab_classes}
\resizebox{\columnwidth}{!}{%
\begin{tabular}{c|cc|cc|cc}
\textbf{}         & \multicolumn{2}{c|}{\textbf{BC}} & \multicolumn{2}{c|}{\textbf{sOPF}} & \multicolumn{2}{c}{\textbf{SVM-RBF}} \\ \cline{2-7} 
                  & \textbf{HC}     	     & \textbf{PD}    	        & \textbf{HC}      	    & \textbf{PD}     	        & \textbf{HC}       	      & \textbf{PD}      \\ \hline
\textbf{dOPF}     & 67.41$\pm$8.36           & \textbf{69.58$\pm$11.78} & 66.67$\pm$9.62            & \textbf{68.33$\pm$12.60}  & \textbf{73.70$\pm$13.36}    & 66.67$\pm$12.20            \\
\textbf{hOPF}    & 67.41$\pm$8.36           & \textbf{69.58$\pm$11.78} & 66.67$\pm$9.62            & \textbf{68.33$\pm$12.60}  & \textbf{83.70$\pm$11.40}    & 64.17$\pm$11.92            \\
\textbf{hOPF-25} & 82.96$\pm$7.99           & 55.00$\pm$13.81          & 83.70$\pm$7.99            & 53.33$\pm$13.95           & \textbf{81.85$\pm$7.11}     & 65.83$\pm$16.68            \\
\textbf{hOPF-50} & \textbf{89.99$\pm$6.01}  & 58.75$\pm$11.52          & \textbf{89.63$\pm$5.50}   & \textbf{57.92$\pm$13.25}  & \textbf{79.63$\pm$13.72}    & \textbf{75.833$\pm$20.44}  \\
\textbf{hOPF-75} & 81.48$\pm$8.31           & \textbf{63.75$\pm$14.01} & 81.11$\pm$9.10            & \textbf{64.17$\pm$14.07}  & \textbf{78.52$\pm$11.67}    & \textbf{82.92$\pm$13.04}
\end{tabular}}
\end{table}

Since the difference between dOPF and hOPF relies on whether the visual words selected in the intermediate layers are used or not in the final dictionary, it must be concluded that the computational load for dictionary learning is the same.

\section{Conclusion and Future Work}
\label{s.conclusion}

This work introduced a hierarchical-learning approach using the Deep Optimum-Path Forest to design visual dictionaries. The proposed approach was assessed and compared against a previous approach proposed by Afonso et al.~\cite{AfonsoSIBGRAPI:17} in the context of Parkinson's disease identification. The experiments used six datasets derived from signal data collected when individuals were submitted to a handwriting exam. The exam is comprised of tasks supposed not to be trivial to Parkinson's disease patients, and the usage of signals allows to detect subtle variations.

The main contributions of this work rely on the introduction of the proposed approach itself, its application in the context of automatic PD detection, and the usage of Restricted Boltzmann Machine for data compression. Experimental results showed the potential of hierarchical-learning approaches where interesting results were achieved. A general analysis pointed improvements in most scenarios and the proposed approach always figured the best results. An in-depth investigation showed a more considerable improvement in accuracy in the healthy individuals class in most scenarios.

With respect to the compressed representations, RBM provided good models and achieved very interesting results (it either outperformed or was statistically similar to the original-sized representation and dOPF) in $12$ out of $18$ configurations (i.e., pair [dictionary learner, classifier] ) for the HC class, and $10$ out of $18$ configurations for the PD class. Regarding future work, we aim to study different ways to create hierarchical representations instead of the concatenation.

\section*{Acknowledgments}
\sloppypar{The authors are grateful to FAPESP grants \#2013/07375-0, \#2014/12236-1 and \#2016/19403-6, Capes, and CNPq grants \#306166/2014-3, \#307066/2017-7, and \#303808/2018-7.}

\bibliography{references}

\end{document}